\pdfoutput=1
\documentclass[11pt,a4paper]{article}
\usepackage[hyperref]{emnlp2018}
\usepackage{bm}
\usepackage{times}
\usepackage{amsmath}
\usepackage{latexsym}
\usepackage{amssymb}
\usepackage{booktabs}
\usepackage{graphicx}
\usepackage[export]{adjustbox}
\usepackage{url}
\usepackage{caption}
\usepackage{subcaption}

\aclfinalcopy 
\def\aclpaperid{236} 

\newcommand\Modelname{Embedded-State Latent CRF}
\newcommand\modelname{embedded-state latent CRF}

\newcommand{\comment}[1]{}

\title{Embedded-State Latent Conditional Random Fields\\ for Sequence Labeling}

\author{Dung Thai \enskip Sree Harsha Ramesh \enskip Shikhar Murty \enskip Luke Vilnis \enskip Andrew McCallum \\
  College of Information and Computer Sciences \\
  University of Massachusetts Amherst\\
  \texttt{\{dthai, shramesh, smurty, luke, mccallum\}@cs.umass.edu}}

\date{}

\begin{document}
\maketitle
\begin{abstract}
Complex textual information extraction tasks are often posed as sequence labeling or \emph{shallow parsing}, where fields are extracted using local labels made consistent through probabilistic inference in a graphical model with constrained transitions. Recently, it has become common to locally parametrize these models using rich features extracted by recurrent neural networks (such as LSTM), while enforcing consistent outputs through a simple linear-chain model, representing Markovian dependencies between successive labels. However, the simple graphical model structure belies the often complex non-local constraints between output labels. For example, many fields, such as a first name, can only occur a fixed number of times, or in the presence of other fields. While RNNs have provided increasingly powerful context-aware local features for sequence tagging, they have yet to be integrated with a global graphical model of similar expressivity in the output distribution. Our model goes beyond the linear chain CRF to incorporate multiple hidden states per output label, but parametrizes their transitions parsimoniously with low-rank log-potential scoring matrices, effectively learning an embedding space for hidden states. This augmented latent space of inference variables complements the rich feature representation of the RNN, and allows exact global inference obeying complex, learned non-local output constraints. We experiment with several datasets and show that the model outperforms baseline CRF+RNN models when global output constraints are necessary at inference-time, and explore the interpretable latent structure.
\end{abstract}

\section{Introduction}


As with many other prediction tasks involving complex structured outputs, such as image segmentation \cite{chen2016deeplab}, machine translation \cite{bahdanau2015neural}, and speech recognition \cite{hinton2012deep}, deep neural networks (DNNs) for sequence labeling and shallow parsing have become standard tools for for information extraction \cite{collobert2011natural,lample2016neural}. In the language of structured prediction, DNNs process the input sequence to produce a rich \emph{local} parametrization for the output prediction model. However, output variables obey a variety of hard and soft constraints --- for example, in sequence tagging tasks such as named entity recognition, I-PER cannot follow B-ORG.

Interestingly, even with such powerful local featurization, the DNN model does not automatically capture a mode of the output distribution through local decisions alone, and can violate these constraints. Successful applications of DNNs to sequence tagging gain from incorporating a simple linear chain probabilistic graphical model to enforce consistent output predictions \cite{collobert2011natural,lample2016neural}, and more generally the addition of a graphical model to enforce output label consistency is common practice for other tasks such as image segmentation \citep{chen2016deeplab}.

Previous work in DNN-featurized sequence tagging with graphical models for information extraction has limited its output structure modeling to these simple local Markovian dependencies. In this work, we explore the addition of latent variables to the prediction model, and through a parsimonious factorized parameter structure, perform representation learning of hidden state embeddings in the graphical model, complementary to the standard practice of representation learning in the local potentials of the segmentation model. By factorizing the log-potentials of the hidden state transition matrices, we are able to learn large numbers of hidden states without overfitting, while the latent dynamics add the capability to learn global constraints on the overall prediction, without sacrificing efficient exact inference. 

While soft and hard global constraints have a rich history in sequence tagging \cite{koo2010dual, rush2012tutorial, anzaroot2014learning}, they have been underexplored in the context of neural-network based feature extraction models. In response, we present a latent-variable CRF model with a novel mechanism for learning latent constraints without overfitting, using low-rank embeddings of large-cardinality latent variables. For example, these non-local constraints appear in fine-grained nested field extraction, which requires hierarchical consistency between the sub-tags of an entity. Further, information extraction and slot filling tasks often require domain specific constraints --- for example, we must avoid extracting the same field multiple times. A good combination of input featurization and output modeling is needed to capture these structural dependencies.

\begin{figure}
\centering
\includegraphics[width=0.5\textwidth]{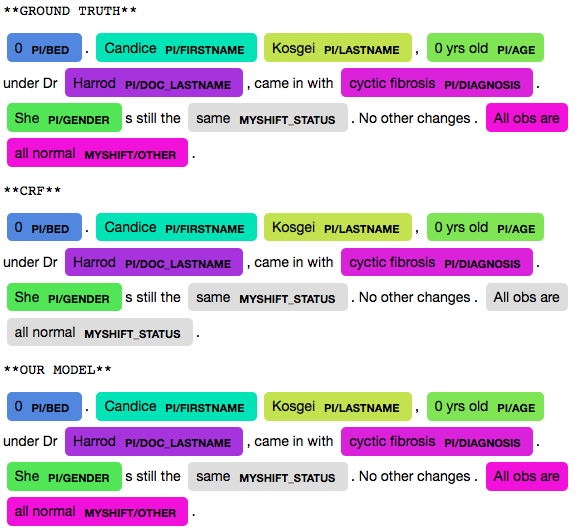}
\caption{An example result from the CLEF eHealth dataset. The soft output constraint suggests tagging patient status as \textit{Myshift/Others} if there already is a \textit{Myshift\_Status} tag. Note that we have the same phrase tagged as \textit{Myshift\_Status} in the training dataset.}
\label{fig:eyecandy}
\end{figure}

In this work we present a method for sequence labeling in which representation learning is applied not only to inputs, but also to output space, in the form of a lightly parameterized transition function between a large number of latent states.  We introduce a hidden state variable and learn the model dynamics in the hidden state space rather than the label state space. This relaxes the Markov assumption between output labels and allows the model to learn global constraints. To avoid the quadratic blowup in parameters with the size of the latent state space, we factorize the transition log-potentials into a low-rank matrix, avoiding overfitting by effectively learning parsimonious embedded representations of the latent states. While the low rank \emph{log}-potential matrix does not improve test-time inference speed, we can perform exact Viterbi inference to compute the labeling sequence. Figure \ref{fig:eyecandy} shows an example where our model finds the correct labeling sequence while a standard DNN+CRF model fails, by obeying a global constraint learned from the training data.

We examine the performance of the \Modelname~on two datasets: citation extraction on the UMass Citations dataset and medical record field extraction on the CLEF dataset. We observe improved performance in both tasks, whose outputs obey complex structural dependencies that are not able to be captured by RNN featurization. Our biggest improvement comes in the medical domain, where the small training set gives our parsimonious approach to output representation learning an extra advantage.

\section{Proposed Model}
\subsection{Problem Formulation}
We consider the sequence labeling task, defined as follows. Given an input text sequence with $T$ tokens $\mathbf{x} = \{x_1, x_2,..., x_T\}$, find a corresponding output sequence $\mathbf{y} = \{y_1, y_2,..., y_T\}$ where each output symbol $y_i$ is one of $N$ possible output labels. There are structural dependencies between the output labels, and resolving such dependencies is necessary for good performance.

\subsection{Background}
The input featurization in our model is similar to previously mentioned existing methods for tagging with DNNs \citep{collobert2011natural}. We represent each input token $x_t$ with a word embedding $w_t$. We then feed the embedded sequence $\mathbf{w} = \{w_1, w_2, ..., w_T\}$ into a bidirectional LSTM \cite{graves2005framewise}. As a result, each input $x_t$ is associated with a contextualized feature vector $f_t = [\overrightarrow{f_t}; \overleftarrow{f_t}] \in \mathbb{R} ^d$ where $\overrightarrow{f_t}$ and $\overleftarrow{f_t}$ represent the left and right context at time step $t$ of the sequence.

In this work, we concern ourselves with the mapping from these input features to a distribution over output label sequences.




\begin{figure}[!ht]
\centering
\captionsetup[figure]{position=b}
\centering
\subcaptionbox{Softmax\label{fig:indep}}{\includegraphics[width=4em]{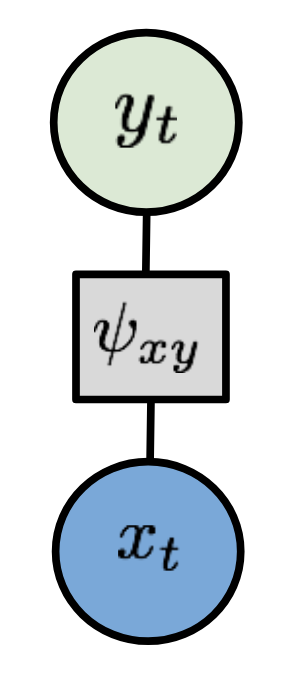}}
\hspace{2.5em}
\vspace{\baselineskip}
\subcaptionbox{Linear-Chain CRF\label{fig:crf}}{\includegraphics[width=8.2em]{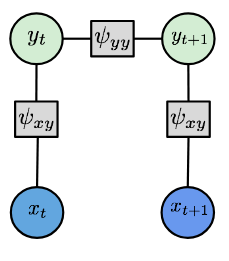}}
\subcaptionbox{\Modelname\label{fig:eslcrf}}{\includegraphics[width=18em]{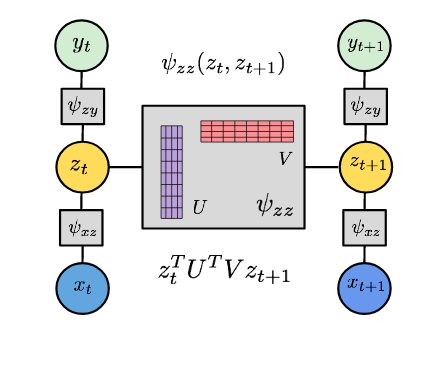}}
\caption{Comparing PGMs for tag prediction.}
\label{fig:pgms}
\end{figure}

A straightforward solution is to use a feed-forward network to map the feature vector to the corresponding label. From a probabilistic perspective, this method is equivalent to the probabilistic graphical model in Fig.\ref{fig:indep}. Here, the goal is to estimate the posterior distribution:
\begin{align}
\mathbb{P} (\mathbf{y} \mid \mathbf{x}) = \prod_{i=1}^{T} P(y_t \mid x_t) = \prod_{i=1}^{T} \psi(y_t; x_t)
\label{eq:fully-factorized}
\end{align}
where the joint distribution over the sequence is fully factorized, i.e. there is no structural dependency between $y_t$ and the distribution $P(y_t \mid x_t)$ is parameterized by a deep neural network $\psi(y_t; x_t) = feed\_forward(f_t,\ y_t)$. This model ignores all the structural dependencies between the output labels during prediction, though not featurization, and has been found unsuitable for structured prediction tasks on sequences \cite{collobert2011natural}. 

In order to enforce some local output consistency, Collobert et al. \shortcite{collobert2011natural} introduce a linear chain Conditional Random Field (CRF) layer to the model (Fig.\ref{fig:pgms}b). They define the energy function for a particular configuration as follows
\begin{align}
\mathcal{E}(\mathbf{y} \mid \mathbf{x}) = \sum_{t=1}^{T} \psi_{xy}(x_t, y_t) + \psi_{yy}(y_t, y_{t+1})
\end{align}
where the local log-potentials $\psi_{xy}$ are parameterized by a DNN, and (for their application) the edge log-potentials $\psi_{yy}$ are parameterized by an input-independent parameter matrix, modeling the intra-state dependencies under a Markovian assumption, giving the data log-likelihood as
\begin{align}
\log \mathbb{P} (\mathbf{y} \mid \mathbf{x}) = \mathcal{E}(\mathbf{y} \mid \mathbf{x})-\log \sum_{\mathbf{y}'} \exp(\mathcal{E}(\mathbf{y}' \mid \mathbf{x}))
\label{eq:log-likelihood}
\end{align} 
\citet{collobert2011natural} show a $+1.71$ performance gain in Named-Entity Recognition (NER) by explicitly enforcing these local structural dependencies. However, the Markov assumption is limiting, and much of the gain  comes from enforcing deterministic hard constraints of the segmentation encoding (e.g. \textit{I-ORG} cannot go after \textit{B-PER}). Similar types of local gains come from hierarchical tagging schemes (e.g. \textit{I-DATE} should be tagged as \textit{I-VENUE/DATE} if it appears inside the \textit{I-VENUE/*} segment). We would like to model, and learn, global, semantically meaningful soft constraints, e.g. \textit{BOOKTITLE} should become \textit{TITLE} if another \textit{TITLE} does not appear in the same citation \cite{anzaroot2014learning}. The state transition dynamics of the linear-chain CRF model are limited by a restriction to interaction between $N$ output labels. The information-rich features $f_t \in \mathbb{R}^d$ input to the local potential are restricted to a local preference over the $N$ labels in output space, failing to exploit the full power of the underlying feature space.

\subsection{\Modelname}
Our proposed model, the \modelname, is shown in Figure \ref{fig:eslcrf}. We introduce a sequence of hidden states $\mathbf{z} = \{z_1, z_2, ..., z_T\}$ where $z_t$ is one of $M$ possible discrete hidden states and $M >> N$. Similarly, the corresponding energy for a particular joint configuration over $\mathbf{y}$ and $\mathbf{z}$ is
\begin{align}
\mathcal{E}(\mathbf{y}, \mathbf{z} \mid \mathbf{x}) = \sum_{t=1}^T & \psi_{xz}(x_t, z_t) + \psi_{yz}(y_t, z_t) \nonumber \\ &+ \psi_{zz}(z_t, z_{t+1}) 
\label{eq:energy}
\end{align}
where $\psi_{xz}(x_t, z_t)$, $\psi_{yz}(y_t, z_t)$ are the local interaction log-potentials between the input features and hidden states, and the hidden states and output states, respectively. The hidden state dynamics come from the log-scores $\psi_{zz}(z_t, z_{t+1})$ for transitioning between hidden state $z_t$ to $z_{t+1}$. The posterior distribution over output labels can be computed by summing over all possible configurations of $\mathbf{z}$
\begin{align}
\mathbb{P}(\mathbf{y} \mid \mathbf{x}) = \frac{1}{Z} \sum_{\mathbf{z}} \exp \left( \mathcal{E}(\mathbf{y}, \mathbf{z} \mid \mathbf{x}) \right)
\label{eq:our-posterior}
\end{align}
where $Z = \sum_{\mathbf{y}'} \sum_{\mathbf{z}'} \exp \left( \mathcal{E}(\mathbf{y}', \mathbf{z}' \mid \mathbf{x}) \right)$ is the partition function.
The local log-potentials $\psi_{xz}(x_t, z_t)$ are produced by an affine transform from the RNN feature extractor, and the output potentials $\psi_{zy}(z_t, y_t)$ are many-to-one mappings from the hidden state, with learned potentials but pre-allocated numbers of states for each output label. 


\textbf{Factorized transition log-potentials} We empirically observe that introducing a large number of hidden states can lead to overfitting, due to over-parameterization of the output dependencies. For example, \textit{JOURNAL} often co-occurs with \textit{PAGES} but \textit{JOURNAL} is not strictly accompanied by \textit{PAGES} \cite{anzaroot2014learning}. Therefore, we regularize the state transition log-potential with a low-rank constraint, forming an embedding matrix wherein state transition interaction scores are mediated through low-dimensional \emph{state embeddings} rather than a fully unconstrained parameter matrix. Instead of learning $A \in \mathbb{R}^{M \times M}$, a full-rank hidden state transition potential, we learn a low-rank model $A = U^{T} V$ where $U$ and $V$ are two \textit{rank-k} matrices. This reduces the number of parameters from $M^2$ to $2Mk$ (where $k << M$) and shares statistical strength when learning transitions between similar states.

\textbf{Inference.} The brute-force computation of the posterior distribution using (\ref{eq:our-posterior}) is intractable, especially with the large number of hidden states. Fortunately, both the energy and the partition function can be computed efficiently using tree belief propagation. Due to the deterministic mapping from hidden states to outputs, we can simply fold the local input \emph{and} output potentials  $\psi_{xz}(x_t, z_t)$ and $\psi_{yz}(y_t, z_t)$ into the edge potentials and perform the forward-backward algorithm as in a standard linear-chain CRF. This deterministic mapping also lets us enforce hard transition constraints while retaining exact inference. Furthermore, since our implementation is in PyTorch \cite{paszke2017automatic}, we only need to implement the forward pass, as automatic differentiation (back-propagation) is equivalent to the backward pass \cite{eisner2016inside}.

\textbf{MAP inference.} At test time, we run the Viterbi algorithm to search for the best configuration over $\mathbf{z}$ rather than over $\mathbf{y}$. Mapping from the hidden state $z_t$ to the output label $y_t$ is deterministic given the output state embedding.

\section{Related Work}

Much deep learning research concerns itself with learning to represent the structure of input space in a way that is highly predictive of the output. In this work, while using state-of-the-art sequence tagging baselines for input representation learning, we concern ourselves with learning the global structure of the output space of label sequences, as well as fine-grained local distinctions in output space. While representation learning in the form of fine-grained, discrete, latent state transitions in the output space has been explored in this context (e.g. various latent-variable conditional random fields \cite{quattoni2007hidden, sutton2007dynamic,morency2007latent} and latent structured support vector machines \cite{yu2009learning}), we enable the use of many more hidden states without overfitting by factorizing the log-potential transition matrices and modeling the log-scores of latent state interactions as products of low-dimensional embeddings, effectively performing feature learning in output space.

A simple linear-chain CRF over the labels was used in early applications of deep learning to sequence tagging \cite{collobert2011natural}, as well as the most recent high-performing segmentation models for named entity recognition \cite{lample2016neural}. Outside of NLP, in tasks such as computer vision, certain classes of fully-connected graphical models over the output pixels have been used for multi-dimensional smoothing \cite{adams2010fast, krahenbuhl2011efficient}, borrowing techniques for the graphics literature.

However, none of these models performs representation learning in the output space, as in the case of our proposed embedded latent-state model. \citet{srikumar2014learning} propose a similar factorized representation of output labels and their transitions, but only apply this to pairwise transitions of output labels and not latent dynamics of the whole sequence, while we believe the biggest gains are to be found by marrying representation learning techniques with latent variable methods.

In the graphical models literature, the most similar work to ours is the Latent-Dynamic CRF of \citet{morency2007latent}, who propose the same graphical model structure, without the deep input featurization, or more importantly, the learned embedded factorization of transition scores. Additionally, that work uses a deterministic mapping of equal numbers of hidden states to output labels, while we have a hard-constrained (hidden states to output variables are always many-to-one), but learned, potential with different outputs pre-allocated different numbers of states based on corpus frequency.

Many graphical models have been proposed for natural language processing under hard and soft global constraints, e.g. \cite{koo2010dual, anzaroot2014learning, vilnis2015bethe}, many based on dual decomposition \cite{rush2012tutorial}. However, the constraints are often fixed, and even when learned \cite{anzaroot2014learning,vilnis2015bethe}, the learning is done simply on constraint weights generated from pre-made templates, the construction of which requires domain knowledge.

Finally, Structured Prediction Energy Networks \citep{belanger2016structured, belanger2017end} have been used for NLP tasks such as semantic role labeling, but they perform approximate inference through gradient descent on a learned energy function over labelings, effectively a fully-connected graphical model, while our model sits more clearly within the framework of graphical models, permitting exact inference with only nonconvex learning, common to all latent-variable models.

\section{Experiments}
We experiment on two datasets with a rich output label space, the UMass Citations dataset \cite{anzaroot2013new} and the CLEF eHealth dataset \cite{suominen2015benchmarking}. Both of the datasets have a hierarchical label space, enforced by hard transition constraints, making this a form of shallow parsing \cite{anzaroot2014learning}, with additional soft constraints in the label space due to the interdependent nature of the fields being extracted.

\subsection{Datasets}

\subsubsection{UMass Citations}
\label{citationdataset}
We experiment with citation field extraction on the UMass Citations dataset \citep{anzaroot2013new}, a collection of 2476 richly labeled citation strings, each tagged in a hierarchical manner, across a set of 38 entities demarcating both coarse-grained labeled segments, such as title, date, authors and venue, as well as fine-grained inner segments where applicable. The data follows a train/dev/test split of 1454, 655 and 367 citations, with 231085 total tokens.  For example, a person's last name could be tagged as \textsc{authors/person/person-last} or \textsc{venue/editor/person/person-last} depending on whether the person is the author of the cited \textsc{title} or an editor of the publication \textsc{venue}. Similarly, year could be tagged as either \textsc{date/year} or \textsc{venue/date/year} depending on whether it is the cited work's publication date or the publication date of the venue of the cited work. 


\subsubsection{CLEF eHealth}

We perform our second set of sequence labeling experiments on the NICTA Synthetic Nursing Handover dataset \citep{suominen2015benchmarking} for clinical information extraction, consisting of 101 documents totaling 32122 tokens.


It is a synthetic dataset of handover records, which contain patient profiles as written by a registered nurse (RN) working in the medical ward and delivering verbal handovers to another nurse at a shift change by the patient’s bedside. A document is typically 100-300 words long, and the included handover information contains five coarse entities i.e, \textsc{PatientIntroduction}, \textsc{MyShift}, \textsc{Appointments}, \textsc{Medication} and \textsc{FutureCare}. Similar to the setup of the citation field extraction task described in Section \ref{citationdataset}, each of these coarse categories has a further level of nested finer labels and the entities to be identified are all hierarchical in nature.  For example, the \textsc{PatientIntroduction} section contains entities such as \textsc{PatientIntroduction/Lastname} and \textsc{PatientIntroduction/UnderDr\_Lastname}, the \textsc{Appointments} section contains \textsc{Appointment/Procedure\_ClinicianLastname}, and \textsc{Medication} contains \textsc{Medication/Dosage} and \textsc{Medication/Medicine}. There are a total of $35$ such fine-grained entities. In addition to the hard-constrained hierarchical structure of the labels, the task also exhibits interesting global constraints, such as only tagging the first occurrence of the patient's gender, or the convention of labeling the most brief description of a nurse's shift status as \textsc{Myshift/Status}, while the details of the shift are tagged as \textsc{Myshift/Other}. In such cases, information extraction benefits from modeling output label dependencies, as we show in the results section. 

\subsection{Training Details}
Our baseline is the BiLSTM+CRF model from \citet{lample2016neural}, employing a bidirectional LSTM with 500 hidden units for input featurization to capture long-range dependencies in the input space. Since we do not focus on input featurization, we do not use character-level embeddings in the baseline model. 

Both the baseline model and our EL-CRF model were implemented in PyTorch. For training our models, we use the hyper-parameter settings from the LSTM+CRF model of \citet{lample2016neural}. Although, we did explore different optimizer techniques to enhance SGD such as Adam \citep{kingma2014adam}, Averaged SGD \citep{polyak1992acceleration} and YellowFin \citep{zhang2017yellowfin}, none of them performed as well as mini-batch SGD with a batch-size of $1$. We also employed gradient clipping to a norm of $5.0$, a learning rate of $0.01$, learning rate decay of $0.05$, dropout with $p=0.5$, and early stopping, tuned on the citation development data. We initialized our word level embeddings using pre-trained 100 dimensional Glove embeddings \citep{pennington2014glove}, which gave better performance on our tasks than the skip-n-gram embeddings \citep{ling2015not} used in the original work of \citet{lample2016neural}. The datasets were pre-processed to zero-replace all occurrences of numbers.  Finally, we experimented with both IOBES and IOB tagging schemes, with IOB demonstrating higher performance on our tasks.


\textbf{Embedding size} We tune the embedding size (rank constraint) for the hidden state matrix $A$, varying from $10$ to $40$, alongside the neural network parameters, and report results when fixing the other hyperparameters and varying embedding size, similar to ablation analysis. Table \ref{table:factorsize} shows the impact of different embedding sizes on the performance of the model. We found that a size of $20$ works best for both datasets, confirming the importance of the rank-constrained log-potential when using large-cardinality hidden variables.

\textbf{Mapping tags to hidden states} We find that the mapping between tags and hidden states greatly influences the performance of the model. We experimented with several heuristics (e.g., individual IOB tag count ratio and entity count ratio), and found that allocating a number of hidden states proportional to the entity count gives us the best performance.

\subsubsection{Evaluation}
We report field-level F1 scores as computed using the \verb|conlleval.pl| script. 


Since the train/validation/test splits were clearly defined for the UMass Citation dataset, we trained the models on the training split, tuned the hyper-parameters on the validation split and report the scores on the test dataset. However, as there were only 101 documents in the CLEF eHealth dataset, we report the Leave-One-Out (LOO) cross-validation F1 scores for this dataset i.e., we trained 101 models each with a different held-out document, merged the respective test outputs, and computed the F1 score on this merged output.

 
\subsection{Results}
Table \ref{f1} shows that overall performance on the UMass Citation dataset using the \modelname~($95.18$) is marginally better than the baseline BiLSTM+CRF model ($95.07$). However, examining the entities with the largest F1 score improvement in Table \ref{citationtop5}, we see that they are mostly within the \textsc{venue} section, which has long-range constraints with other sections, giving evidence of the model's ability to learn constraints from the citation dataset. 

\begin{table}[!htbp]
\centering
\begin{tabular}{@{}c|c|c|c@{}}
\textbf{Dataset} & \textbf{CRF} & \textbf{EL-CRF} & \textbf{+} \\ \midrule
\textsc{UMass Citation}   & 95.07        & 95.18  & \textbf{0.11}       \\ \midrule
\textsc{CLEF eHealth}     & 68.66        & 70.32  & \textbf{1.66}       \\ 
\end{tabular}
\caption{Entity-level F1 scores of the \modelname~ and BiLSTM+CRF baseline.}
\label{f1}
\end{table}

\begin{table}[!htbp]
\small  
\centering
\begin{tabular}{@{}c|c|c|c|c@{}}
\textbf{Label}                                                                & \textbf{CRF} & \textbf{EL-CRF} & \textbf{+} & \textbf{S}  \\ \midrule
\textsc{v/department}                                                              & 66.67        & 100             & \textbf{33.33}    & 1  \\ \midrule
\textsc{v/status}                                                                  & 77.78        & 87.5            & \textbf{9.72}      & 9 \\ \midrule
\begin{tabular}[c]{@{}c@{}}\textsc{v/e/p/}\\ \textsc{person\_middle}\end{tabular} & 83.33        & 91.67           & \textbf{8.34}  & 11     \\ \midrule
\textsc{reference\_id}                                                                & 85.11        & 93.02           & \textbf{7.91}      & 22  \\ \midrule
\textsc{v/series}                                                                  & 55.17        & 61.54           & \textbf{6.37}      & 12 \\ \midrule
\textsc{v/address}                                                                 & 78.85        & 84.31           & \textbf{5.46}   & 46   
\end{tabular}
\caption{Top 5 entities in terms of F1 improvement on the UMass Citation Dataset. The column \textit{S} shows the support for a given entity in the test dataset. Key for contracted entity names: \textsc{v}: \textsc{venue}, \textsc{e}: \textsc{editor}, \textsc{p}: \textsc{person}}
\label{citationtop5}
\end{table}

Table \ref{f1} demonstrates that EL-CRF outperforms the BiLSTM+CRF on both datasets, with larger gains on the much smaller CLEF data. Table \ref{cleftop5} shows the top-gaining entities include \textsc{Medication\_Medicine} and \textsc{Medication\_Dosage}, due to the global constraint that those entities always co-occur. 

\begin{table}[]
\small 
\centering
\begin{tabular}{@{}c|c|c|c|c@{}}
\textbf{Label}                                                                                 & \textbf{CRF} & \textbf{EL-CRF} & \textbf{+} & \textbf{S} \\ \midrule
\begin{tabular}[c]{@{}c@{}}\textsc{P/Dr/GivenNames/}\\ \textsc{Initials}\end{tabular} & 33.33        & 64.29          & \textbf{30.96}    & 15  \\ \midrule
\begin{tabular}[c]{@{}c@{}}\textsc{A/Procedure/}\\ \textsc{Time}\end{tabular}                          & 34.78        & 53.66           & \textbf{18.88} & 28     \\ \midrule
\textsc{M/Medicine}                                                                           & 55.28        & 71.1            & \textbf{15.82}    & 157  \\ \midrule
\begin{tabular}[c]{@{}c@{}}\textsc{FA/Warning/}\\ \textsc{AbnormalResult}\end{tabular}               & 0            & 11.43           & \textbf{11.43}   & 59   \\ \midrule
\textsc{M/Dosage}                                                                             & 9.09         & 18.75           & \textbf{9.66}   & 37   
\end{tabular}
\caption{Top 5 entities in terms of F1 improvement on the CLEF eHealth dataset. Key for contracted entity names: \textsc{p/dr}: \textsc{patient\_introduction/under\_dr}, \textsc{a}: \textsc{appointment}, \textsc{m}: \textsc{medication}, \textsc{fa}: \textsc{future\_alert}}
\label{cleftop5}
\end{table}

\begin{table}[]
\small 
\centering
\begin{tabular}{@{}c|c|c@{}}
\textbf{Factor Size}                                          & \textbf{UMass Citation} & \textbf{CLEF eHealth} \\ \midrule
10                                                            & 94.92                   & 70.06                 \\ \midrule
\textbf{20}                                                            & \textbf{95.18}          & \textbf{71.51}        \\ \midrule
30                                                            & 94.91                   & 69.92                 \\ \midrule
40                                                            & 94.88                   & 70.33                 \\ \midrule
\begin{tabular}[c]{@{}c@{}}Full Rank\end{tabular} & 95.13                   & 71.11                
\end{tabular}
\caption{Comparison of F1 scores obtained by varying the \textit{factor\_size} parameter, and setting the other model and neural network parameters from the model with the best cross validation.}
\label{table:factorsize}
\end{table}

\section{Qualitative Analysis}
In this section, we provide qualitative evidence that the \modelname~ learns constraints which are not captured by the standard CRF. 

\begin{figure}[!ht]
\centering
\includegraphics[width=20em]{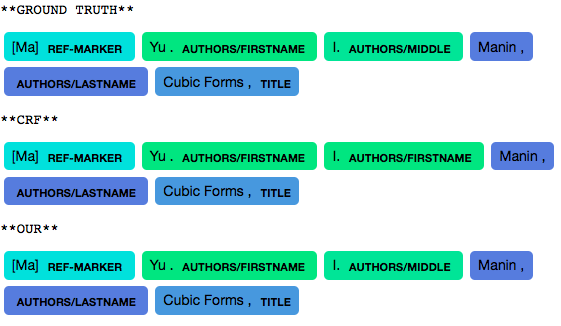}
\caption{Authors name constraint violation}
\label{fig:ex-firstname}
\end{figure}

First, we pick a few representative examples from the UMass Citations dataset and discuss when our model is able to correctly determine the label sequence based on the output constraints. In addition to the the hard constraints arising from hierarchical segmentation, this dataset also exhibits empirical pairwise constraints between fields e.g. two different authors' first names cannot be placed next to each other. Figure \ref{fig:ex-firstname} demonstrates that the CRF model fails to enforce such constraints. 

\begin{figure}[!ht]
\centering
\includegraphics[width=20em]{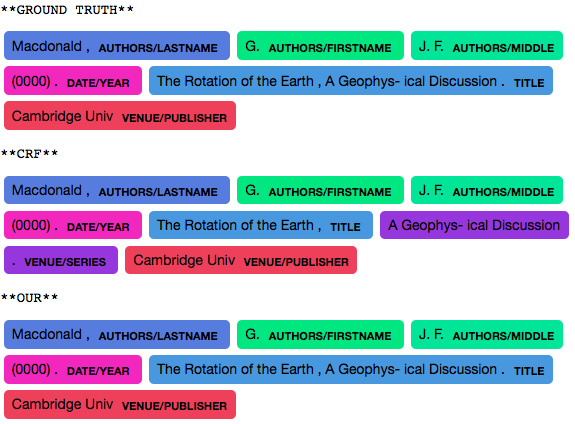}
\caption{Title should not co-occur with series.}
\label{fig:ex-title-series}
\end{figure}

\noindent Another constraint we observe in the citation data is that the \textit{Venue/Series} tag only appears once per citation if \textit{Venue/Booktitle} is also present. Our model obeys this constraint and marks the whole span as \textit{Title} instead of breaking it into \textit{Title} and \textit{Venue/Series}, even though the input text for that segment in isolation could represent a valid series (Figure\ref{fig:ex-title-series}). 

\begin{figure}[!ht]
\centering
\includegraphics[width=20em]{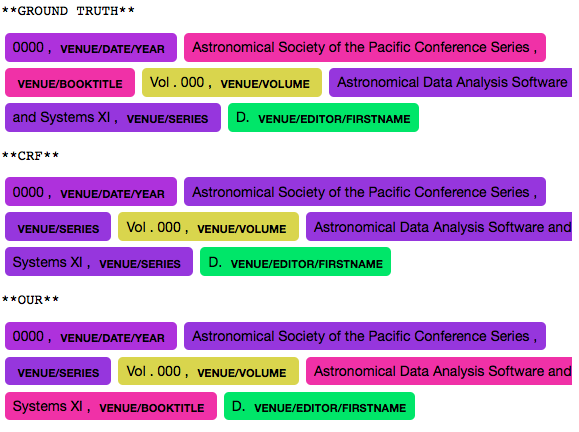}
\caption{There is at most one series per citation.}
\label{fig:ex-series}
\end{figure}

\noindent Sometimes output structural dependencies are not able to resolve ambiguity in the labeling sequence. In Figure \ref{fig:ex-series} our model correctly predicts the presence of a \textit{Venue/Booktitle} and a \textit{Venue/Series}, but it fails to correctly assign the entity labels.

The CLEF eHealth dataset holds a different set of constraints than the citation data, and its input sequences are not strong local indicators of the labeling sequence. Therefore, our model shows stronger performance over the Markovian baseline for this dataset. Some of the constraints concern the number of entities per document. For example, we only tag the first occurrence of a gender indicator e.g. \emph{he}, \emph{she}, \emph{her}, etc., or the most general status of a nurse's shift.
\begin{figure}[!ht]
\centering
\includegraphics[width=20em]{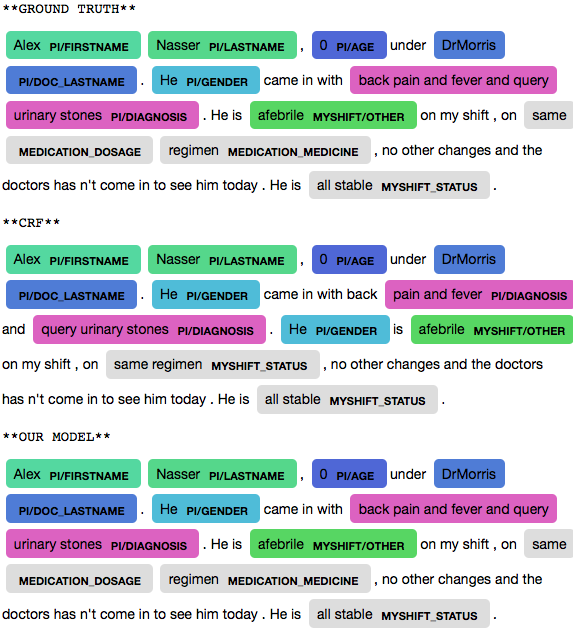}
\caption{The gender indicator constraint and nurse's shift status constraint in the CLEF eHealth dataset.}
\label{fig:ex-clef}
\end{figure}







\begin{figure}[!ht]
\centering
\includegraphics[width=20em]{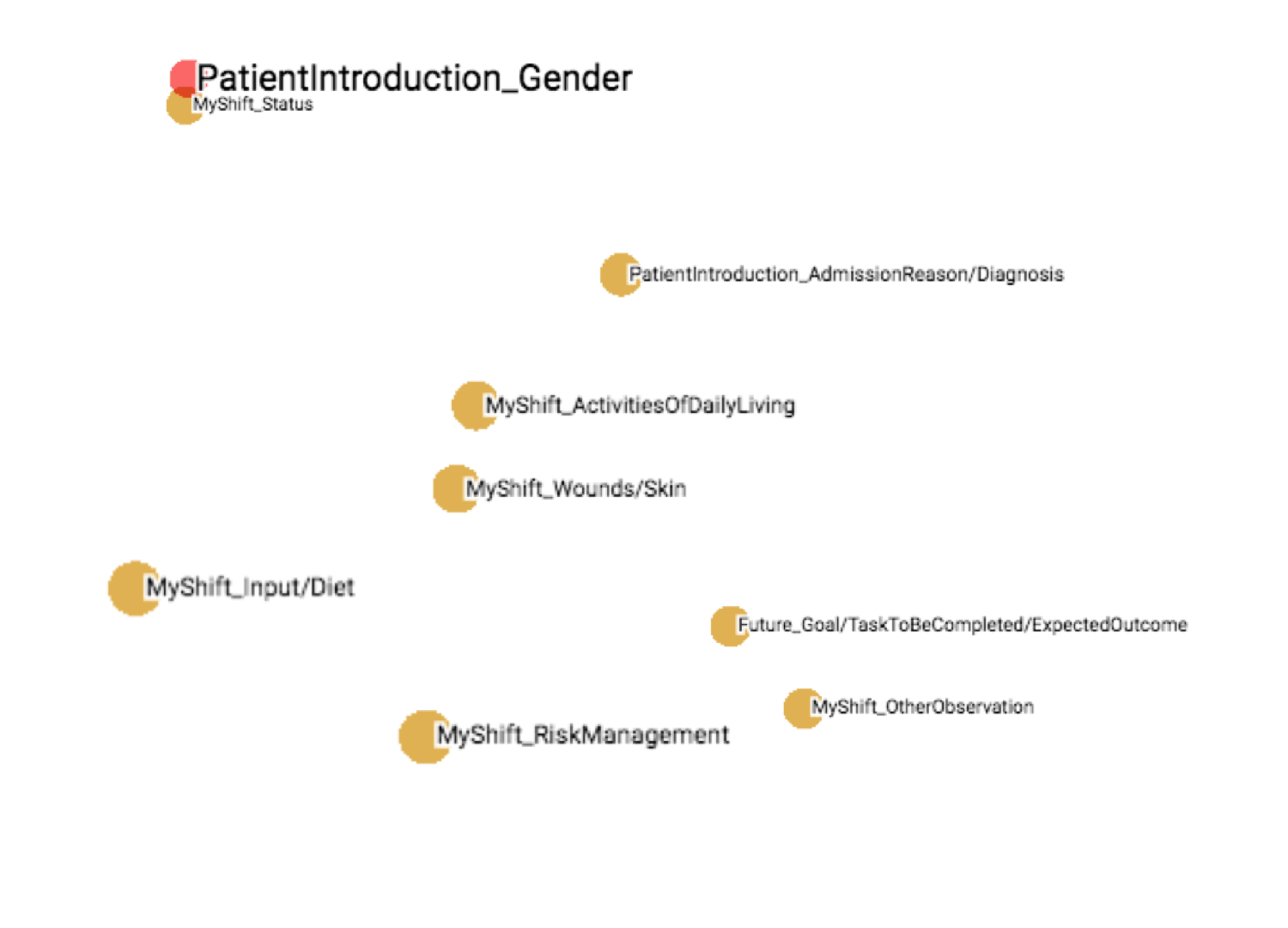}
\caption{A part of the T-SNE clustering of the tag embedding from the CLEF eHealth dataset. The two tag \textsc{PatientIntroduction\_Gender} and \textsc{MyShift\_Status} are under the similar constraint of being tagged only once per document.}
\label{fig:t-sne}
\end{figure}

Finally, a T-SNE \cite{maaten2008visualizing} clustering on the embedding vectors of the output tags, shown in Figure \ref{fig:t-sne}, demonstrates that output structural dependencies can be reflected in tag embedding space.

\section{Conclusion \& Future Work}
We present a latent variable model which not only parametrizes local potentials with the learned features from a deep neural network, but learns embedded representations in a large hidden state space, leveraging feature learning in both the input and output representations. Experimental results demonstrate that our model can learn global structural dependencies in the presence of ambiguities that cannot be resolved by local featurization of the input sequence. We find interpretable structure in the output state embeddings. 

Future work will apply our model to larger datasets with more complex dependencies, and introduce multiple latent states per time-step, enabling exponentially more expressivity in output states at the cost of exact inference. We will also explore approximate inference methods, such as expectation propagation, to speed up message passing in the regime of low-rank log-potentials.
\bibliography{emnlp2018}

\begin{thebibliography}{29}
\expandafter\ifx\csname natexlab\endcsname\relax\def\natexlab#1{#1}\fi

\bibitem[{Adams et~al.(2010)Adams, Baek, and Davis}]{adams2010fast}
Andrew Adams, Jongmin Baek, and Myers~Abraham Davis. 2010.
\newblock Fast high-dimensional filtering using the permutohedral lattice.
\newblock In \emph{Computer Graphics Forum}, volume~29, pages 753--762. Wiley
  Online Library.

\bibitem[{Anzaroot and McCallum(2013)}]{anzaroot2013new}
Sam Anzaroot and Andrew McCallum. 2013.
\newblock A new dataset for fine-grained citation field extraction.

\bibitem[{Anzaroot et~al.(2014)Anzaroot, Passos, Belanger, and
  McCallum}]{anzaroot2014learning}
Sam Anzaroot, Alexandre Passos, David Belanger, and Andrew McCallum. 2014.
\newblock Learning soft linear constraints with application to citation field
  extraction.
\newblock \emph{arXiv preprint arXiv:1403.1349}.

\bibitem[{Bahdanau et~al.(2015)Bahdanau, Cho, and Bengio}]{bahdanau2015neural}
Dzmitry Bahdanau, Kyunghyun Cho, and Yoshua Bengio. 2015.
\newblock Neural machine translation by jointly learning to align and
  translate.
\newblock \emph{ICLR}.

\bibitem[{Belanger and McCallum(2016)}]{belanger2016structured}
David Belanger and Andrew McCallum. 2016.
\newblock Structured prediction energy networks.
\newblock In \emph{International Conference on Machine Learning}, pages
  983--992.

\bibitem[{Belanger et~al.(2017)Belanger, Yang, and McCallum}]{belanger2017end}
David Belanger, Bishan Yang, and Andrew McCallum. 2017.
\newblock End-to-end learning for structured prediction energy networks.
\newblock \emph{ICML}.

\bibitem[{Chen et~al.(2018)Chen, Papandreou, Kokkinos, Murphy, and
  Yuille}]{chen2016deeplab}
Liang-Chieh Chen, George Papandreou, Iasonas Kokkinos, Kevin Murphy, and Alan~L
  Yuille. 2018.
\newblock Deeplab: Semantic image segmentation with deep convolutional nets,
  atrous convolution, and fully connected crfs.
\newblock \emph{PAMI}.

\bibitem[{Collobert et~al.(2011)Collobert, Weston, Bottou, Karlen, Kavukcuoglu,
  and Kuksa}]{collobert2011natural}
Ronan Collobert, Jason Weston, L{\'e}on Bottou, Michael Karlen, Koray
  Kavukcuoglu, and Pavel Kuksa. 2011.
\newblock Natural language processing (almost) from scratch.
\newblock \emph{Journal of Machine Learning Research}, 12(Aug):2493--2537.

\bibitem[{Eisner(2016)}]{eisner2016inside}
Jason Eisner. 2016.
\newblock Inside-outside and forward-backward algorithms are just backprop
  (tutorial paper).
\newblock In \emph{Proceedings of the Workshop on Structured Prediction for
  NLP}, pages 1--17.

\bibitem[{Graves and Schmidhuber(2005)}]{graves2005framewise}
Alex Graves and J{\"u}rgen Schmidhuber. 2005.
\newblock Framewise phoneme classification with bidirectional lstm networks.
\newblock In \emph{Neural Networks, 2005. IJCNN'05. Proceedings. 2005 IEEE
  International Joint Conference on}, volume~4, pages 2047--2052. IEEE.

\bibitem[{Hinton et~al.(2012)Hinton, Deng, Yu, Dahl, Mohamed, Jaitly, Senior,
  Vanhoucke, Nguyen, Sainath et~al.}]{hinton2012deep}
Geoffrey Hinton, Li~Deng, Dong Yu, George~E Dahl, Abdel-rahman Mohamed, Navdeep
  Jaitly, Andrew Senior, Vincent Vanhoucke, Patrick Nguyen, Tara~N Sainath,
  et~al. 2012.
\newblock Deep neural networks for acoustic modeling in speech recognition: The
  shared views of four research groups.
\newblock \emph{IEEE Signal Processing Magazine}, 29(6):82--97.

\bibitem[{Kingma and Ba(2015)}]{kingma2014adam}
Diederik~P Kingma and Jimmy Ba. 2015.
\newblock Adam: A method for stochastic optimization.
\newblock \emph{ICLR}.

\bibitem[{Koo et~al.(2010)Koo, Rush, Collins, Jaakkola, and
  Sontag}]{koo2010dual}
Terry Koo, Alexander~M Rush, Michael Collins, Tommi Jaakkola, and David Sontag.
  2010.
\newblock Dual decomposition for parsing with non-projective head automata.
\newblock In \emph{EMNLP}.

\bibitem[{Kr{\"a}henb{\"u}hl and Koltun(2011)}]{krahenbuhl2011efficient}
Philipp Kr{\"a}henb{\"u}hl and Vladlen Koltun. 2011.
\newblock Efficient inference in fully connected crfs with gaussian edge
  potentials.
\newblock In \emph{Advances in neural information processing systems}, pages
  109--117.

\bibitem[{Lample et~al.(2016)Lample, Ballesteros, Subramanian, Kawakami, and
  Dyer}]{lample2016neural}
Guillaume Lample, Miguel Ballesteros, Sandeep Subramanian, Kazuya Kawakami, and
  Chris Dyer. 2016.
\newblock Neural architectures for named entity recognition.
\newblock \emph{ACL}.

\bibitem[{Ling et~al.(2015)Ling, Tsvetkov, Amir, Fermandez, Dyer, Black,
  Trancoso, and Lin}]{ling2015not}
Wang Ling, Yulia Tsvetkov, Silvio Amir, Ramon Fermandez, Chris Dyer, Alan~W
  Black, Isabel Trancoso, and Chu-Cheng Lin. 2015.
\newblock Not all contexts are created equal: Better word representations with
  variable attention.
\newblock In \emph{Proceedings of the 2015 Conference on Empirical Methods in
  Natural Language Processing}, pages 1367--1372.

\bibitem[{Maaten and Hinton(2008)}]{maaten2008visualizing}
Laurens van~der Maaten and Geoffrey Hinton. 2008.
\newblock Visualizing data using t-sne.
\newblock \emph{Journal of machine learning research}, 9(Nov):2579--2605.

\bibitem[{Morency et~al.(2007)Morency, Quattoni, and
  Darrell}]{morency2007latent}
Louis-Philippe Morency, Ariadna Quattoni, and Trevor Darrell. 2007.
\newblock Latent-dynamic discriminative models for continuous gesture
  recognition.
\newblock In \emph{Computer Vision and Pattern Recognition, 2007. CVPR'07. IEEE
  Conference on}, pages 1--8. IEEE.

\bibitem[{Paszke et~al.(2017)Paszke, Gross, Chintala, Chanan, Yang, DeVito,
  Lin, Desmaison, Antiga, and Lerer}]{paszke2017automatic}
Adam Paszke, Sam Gross, Soumith Chintala, Gregory Chanan, Edward Yang, Zachary
  DeVito, Zeming Lin, Alban Desmaison, Luca Antiga, and Adam Lerer. 2017.
\newblock Automatic differentiation in pytorch.
\newblock In \emph{NIPS-W}.

\bibitem[{Pennington et~al.(2014)Pennington, Socher, and
  Manning}]{pennington2014glove}
Jeffrey Pennington, Richard Socher, and Christopher Manning. 2014.
\newblock Glove: Global vectors for word representation.
\newblock In \emph{Proceedings of the 2014 conference on empirical methods in
  natural language processing (EMNLP)}, pages 1532--1543.

\bibitem[{Polyak and Juditsky(1992)}]{polyak1992acceleration}
Boris~T Polyak and Anatoli~B Juditsky. 1992.
\newblock Acceleration of stochastic approximation by averaging.
\newblock \emph{SIAM Journal on Control and Optimization}, 30(4):838--855.

\bibitem[{Quattoni et~al.(2007)Quattoni, Wang, Morency, Collins, and
  Darrell}]{quattoni2007hidden}
Ariadna Quattoni, Sybor Wang, Louis-Philippe Morency, Morency Collins, and
  Trevor Darrell. 2007.
\newblock Hidden conditional random fields.
\newblock \emph{IEEE transactions on pattern analysis and machine
  intelligence}, 29(10).

\bibitem[{Rush and Collins(2012)}]{rush2012tutorial}
Alexander~M Rush and MJ~Collins. 2012.
\newblock A tutorial on dual decomposition and lagrangian relaxation for
  inference in natural language processing.
\newblock \emph{Journal of Artificial Intelligence Research}, 45:305--362.

\bibitem[{Srikumar and Manning(2014)}]{srikumar2014learning}
Vivek Srikumar and Christopher~D Manning. 2014.
\newblock Learning distributed representations for structured output
  prediction.
\newblock In \emph{Advances in Neural Information Processing Systems}, pages
  3266--3274.

\bibitem[{Suominen et~al.(2015)Suominen, Zhou, Hanlen, and
  Ferraro}]{suominen2015benchmarking}
Hanna Suominen, Liyuan Zhou, Leif Hanlen, and Gabriela Ferraro. 2015.
\newblock Benchmarking clinical speech recognition and information extraction:
  new data, methods, and evaluations.
\newblock \emph{JMIR medical informatics}, 3(2).

\bibitem[{Sutton et~al.(2007)Sutton, McCallum, and
  Rohanimanesh}]{sutton2007dynamic}
Charles Sutton, Andrew McCallum, and Khashayar Rohanimanesh. 2007.
\newblock Dynamic conditional random fields: Factorized probabilistic models
  for labeling and segmenting sequence data.
\newblock \emph{Journal of Machine Learning Research}, 8(Mar):693--723.

\bibitem[{Vilnis et~al.(2015)Vilnis, Belanger, Sheldon, and
  McCallum}]{vilnis2015bethe}
Luke Vilnis, David Belanger, Daniel Sheldon, and Andrew McCallum. 2015.
\newblock Bethe projections for non-local inference.
\newblock \emph{arXiv preprint arXiv:1503.01397}.

\bibitem[{Yu and Joachims(2009)}]{yu2009learning}
Chun-Nam~John Yu and Thorsten Joachims. 2009.
\newblock Learning structural svms with latent variables.
\newblock In \emph{Proceedings of the 26th annual international conference on
  machine learning}, pages 1169--1176. ACM.

\bibitem[{Zhang et~al.(2017)Zhang, Mitliagkas, and R{\'e}}]{zhang2017yellowfin}
Jian Zhang, Ioannis Mitliagkas, and Christopher R{\'e}. 2017.
\newblock Yellowfin and the art of momentum tuning.
\newblock \emph{arXiv preprint arXiv:1706.03471}.

\end{thebibliography}
\bibliographystyle{acl_natbib_nourl}



\end{document}